\documentclass[11pt]{article}
\usepackage[margin=1in]{geometry}
\usepackage[T1]{fontenc}
\usepackage{lmodern}
\usepackage{microtype}
\usepackage{graphicx}
\usepackage{booktabs}
\usepackage{amsmath,amssymb,amsthm,bm}
\usepackage{multirow}
\usepackage{algorithm}
\usepackage{algorithmic}
\usepackage{siunitx}
\usepackage{xcolor}
\usepackage{enumitem}
\usepackage{caption}
\usepackage{subcaption}
\usepackage{placeins}
\usepackage{array}
\usepackage{natbib}
\usepackage{xurl}
\usepackage{hyperref}
\usepackage[nameinlink,noabbrev]{cleveref}

\newcommand{\PPL}{\mathrm{PPL}}

\let\originalincludegraphics\includegraphics
\renewcommand{\includegraphics}[2][]{%
  \IfFileExists{#2}{\originalincludegraphics[#1]{#2}}{%
    \fbox{\parbox[c][1.0in][c]{0.83\linewidth}{\centering\textbf{Missing figure asset}\\\texttt{\detokenize{#2}}}}%
  }%
}
\title{Orthogonally Mixed Sign Coding with Progressive Residual Refinement for Transformer Weight Compression}

\author{
    Anindya Roy Chowdhury\\
    Los Alamos National Laboratory\\
    \texttt{achowdhury@lanl.gov}
    \and
    Atanu Barai\\
    Los Alamos National Laboratory\\
    \texttt{abarai@lanl.gov}
    \and
    Nandakishore Santhi\\
    Los Alamos National Laboratory\\
    \texttt{nsanthi@lanl.gov}
}

\date{}
\begin{document}
\maketitle
\begin{abstract}
\begin{abstract}
LLM quantization reduces storage but often degrades predictive fidelity at low bit rates. We propose a training-free, CDF-free, Gaussian-CLT-motivated progressive residual-sign quantization framework that uses randomized orthogonal mixing to promote Gaussian-like residual distributions and improve sign-based reconstruction of Transformer weights. Each stage independently normalizes residuals and analytically calibrates reconstruction, progressively recovering quantization errors. The framework enables flexible storage–fidelity trade-offs without learned codebooks, empirical CDFs, or trellis optimization. We evaluate GPT-2 Large and Llama-3-8B across attention and MLP projections. At five residual stages, joint quantization preserves near-baseline predictive fidelity, achieving perplexities of 32.16 and 12.73, compared with full-precision values of 29.89 and 11.07, respectively. Ablations further demonstrate that orthogonal mixing substantially improves reconstruction over identity-based sign coding, while activation-aware scaling improves output fidelity and reduced-precision statistics lower storage overhead with negligible perplexity degradation. Comparisons with conventional quantizers establish the method's progressive fidelity recovery and characterize its storage–quality and runtime trade-offs. These results demonstrate the effectiveness of orthogonally mixed residual-sign coding for flexible, training-free Transformer weight compression.
\end{abstract}

\end{abstract}
\noindent\textbf{Keywords:} large language models, post-training quantization, orthogonal mixing, progressive residual coding.
\section{Introduction}
\label{sec:introduction}

Transformer models \cite{vaswani2017attention}, including GPT-2 and the LLaMA family \cite{radford2019language,touvron2023llama,touvron2023llama2}, impose substantial weight-storage, transfer, and inference-memory costs. Post-training quantization (PTQ) reduces these costs without retraining. Established methods include outlier-aware LLM.int8() \cite{dettmerslambda2022llm}, second-order GPTQ \cite{frantar2022gptq}, activation-aware SmoothQuant and AWQ \cite{xiao2023smoothquant,lin2024awq}, and rotation-based QuaRot and SpinQuant \cite{ashkboos2024quarot,liu2025spinquant}. Yet scalar quantization does not directly exploit relationships among coordinates of a weight vector.

Vector quantization (VQ) captures such structure through joint encoding and has a long history in information theory and signal processing where given a matrix \(W\in\mathbb R^{d\times n}\) of n vectors in dimension d, is mapped to a discrete, finite subset of representative vectors called a codebook. Then, using just the codebook approximately reconstruct the original matrix. Recently, VQ has been central to AI infrastructure with techniques that employ additive codebooks~\cite{egiazarian2024aqlm}, data-aware codebooks and error compensation~\cite{vanbaalen2024gptvq}, structured lattice codebooks~\cite{tseng2024quipsharp}, or trellis coding~\cite{tseng2024qtip}. These methods motivate a question: can vector weights be compressed progressively, with explicit control over reconstruction fidelity, without learned codebooks, trellis search, or distribution-dependent Cumulative Distribution Function(CDF) computations? This question matters because a fixed low-bit representation may offer insufficient fidelity, while storing additional correction stages could recover accuracy at an added storage cost.

In this paper, we introduce a codebook-free, CDF-free, information theoretically near-optimal, \emph{Progressive Residual-sign Vector Codec} where each stage centers and RMS-normalizes a weight or residual vector, applies seeded orthogonal mixing, and encodes the transformed coordinates by their signs. Compact statistics and a fitted least-squares gain support reconstruction; blockwise distortion selects among deterministic transform candidates. The codec then encodes the remaining error as successive sign-coded corrections, accepting a stage only when it decreases blockwise RMS distortion. Any accepted prefix can be decoded, making residual-stage depth an explicit storage--fidelity control rather than requiring a separately trained quantizer for each operating point. Gaussian central-limit behavior motivates mixing but is neither assumed for correctness nor evaluated through CDF or inverse-CDF operations. Activation calibration does not update model weights; no gradients, distillation, or recovery fine-tuning are required.

We evaluate GPT-2 Large and Llama-3-8B under attention-only, MLP-only, and joint attention--MLP quantization, measuring reconstruction error, perplexity, output fidelity, serialized storage, and runtime across residual depths and codec ablations. Our contributions are: (i) a CDF-free, sign-based vector codec with compact reconstruction statistics; (ii) distortion-aware transform selection and monotonic residual refinement; and (iii) a model-level characterization of progressive storage--fidelity trade-offs without retraining.

Our results show that deeper residual refinement substantially improves model fidelity, although it increases storage and reconstruction cost. Under joint attention--MLP quantization at five residual stages, perplexity reaches $32.16$ on GPT-2 Large and $12.73$ on Llama-3-8B, compared with full-precision baselines of $29.89$ and $11.07$, respectively. Ablations identify activation-aware scaling and transform selection as useful contributors to fidelity. These findings demonstrate the potential of training-free progressive residual-sign quantization to balance compression efficiency and model fidelity through adjustable residual-stage depth.

\section{Background and Related Work}
\label{sec:related_work}

Aggressive weight compression requires preserving useful model information within a limited storage budget, including the auxiliary information needed for reconstruction. The choice of representation is therefore central: it determines what information is retained, how weights are reconstructed, and how additional bits improve the approximation. Our work investigates a progressively refinable representation based on orthogonally mixed sign vectors. Its distinguishing contribution is the design of each residual stage, which combines distortion-based selection with reconstruction statistics and transform metadata.

Researchers have proposed different PTQ techniques for compressing Transformer weight matrices \(W\in\mathbb R^{d_{\mathrm{out}}\times d_{\mathrm{in}}}\) without retraining. GPTQ uses approximate second-order reconstruction information \cite{frantar2022gptq}; AWQ protects activation-salient channels \cite{lin2024awq}; SmoothQuant shifts quantization difficulty between weights and activations \cite{xiao2023smoothquant}; and LLM.int8() treats outliers separately \cite{dettmerslambda2022llm}. In spite of their significant contribution these methods don't explore VQ.

Whiles the scalar quantizers encode coefficients, vector quantizers encode blocks. AQLM uses additive codebooks \cite{egiazarian2024aqlm}, and GPTVQ uses data-aware vector coding \cite{vanbaalen2024gptvq}. QuIP\# combines incoherence processing with structured lattice codebooks \cite{tseng2024quipsharp}, while QTIP uses trellis-coded quantization to access high effective vector dimensions without an exponential codebook \cite{tseng2024qtip}. Our codec instead stores transformed signs, reconstruction statistics, and transform metadata, providing a block representation that requires neither centroid lookup nor trellis optimization.

Orthogonal transforms preserve Euclidean norm while redistributing
coordinate magnitudes. QuaRot and SpinQuant use rotations to improve
quantization robustness \cite{ashkboos2024quarot,liu2025spinquant}.
TurboQuant combines random rotations with distribution-matched scalar
quantization and, for inner-product estimation, a one-bit residual
correction \cite{zandieh2026turboquant}.
We directly sign-code orthogonally mixed vectors and recursively
encode their residuals. Our encoder is \emph{CDF-free}: it requires
neither empirical CDF estimation nor inverse-CDF tables.

Residual quantization constructs
\(\widehat x^{(r)}=\sum_{k=1}^{r}\widehat e^{(k)}\), with
\(e^{(k)}=e^{(k-1)}-\widehat e^{(k)}\) and \(e^{(0)}=x\).
RRQ applies low-bit scalar quantizers to a base and its residuals,
producing multi-precision LLM weights from one checkpoint
\cite{luo2026rrq}.
Our contribution is the \emph{residual-stage codec}: each correction
is a distortion-selected, orthogonally mixed sign vector reconstructed
from \((\mu,\sigma,\gamma)\) and transform metadata.
Thus, the distinction lies in the stage representation, rather than
residual refinement itself: additional stages refine the approximation
through sign-vector geometry rather than increased scalar precision.

LoRA and QLoRA adapt model weights through optimization
\cite{hu2021lora,dettmers2023qlora}.
Our quantizer requires neither model retraining nor recovery-adapter
training; calibration examples supply only activation statistics.
This avoids an additional training phase while retaining
distortion-based selection during encoding.

Random orthogonal mixing forms weighted sums of normalized block elements. A central-limit argument suggests approximately Gaussian marginals under suitable regularity conditions, but does not guarantee Gaussianity for arbitrary weights. Gaussianity motivates mixing; the codec itself requires neither Gaussian probability evaluation nor empirical CDF mappings.

\section{Problem Formulation}
\label{sec:problem}

Let us assume that \(\mathcal W_{\mathcal S}=\{W_1,\ldots,W_L\}\) denote the weights selected under quantization scope \(\mathcal S\in\{\mathrm{Attention}, \mathrm{MLP}, \mathrm{Attention+MLP}\}\), and let \(Q_r(\cdot)\) represent progressive residual-sign quantization with \(r\) stages.
\[
\boxed{\widehat W_{\mathrm Q,\ell}^{(r)}=Q_r(W_\ell),}
\]
while untargeted parameters remain pretrained. Parameter distortion is
\[
D_{\mathrm{param}}^{(r,\mathcal S)}
=
\frac{\sum_{\ell\in\mathcal S}
\|W_\ell-\widehat W_{\mathrm Q,\ell}^{(r)}\|_F^2}
{\sum_{\ell\in\mathcal S}|W_\ell|},
\]
with RMSE, relative error, and cosine similarity as complementary measures.

For input \(X\), let \(Z_{\mathrm{FP}}=f_{\mathrm{FP}}(X)\) and
\(Z_{r,\mathcal S}=f_{r,\mathcal S}(X)\). Functional distortion is
measured over valid, unshifted token positions:
\begin{equation}
\label{eq:masked_logit_mse}
\begin{aligned}
D_{\mathrm{logit}}^{(r,\mathcal S)}
&=\frac{\displaystyle\sum_{i,t,v}a_{i,t}
\bigl(Z_{\mathrm{FP},i,t,v}
-Z_{r,\mathcal S,i,t,v}\bigr)^2}
{\displaystyle V\sum_{i,t}a_{i,t}},
\end{aligned}
\end{equation}
where \(a_{i,t}\in\{0,1\}\) masks valid logit positions, and \(V\)
is the vocabulary size. Unlike the next-token target mask below,
\(a_{i,t}\) applies to unshifted token positions. Logit cosine
similarity is also reported. Language-model quality is measured by masked next-token perplexity,
\begin{equation}
\boxed{\PPL=\exp\!\left(
-\frac{\displaystyle\sum_{i=1}^{N}\sum_{t=1}^{T_i-1}
 m_{i,t}\log p(x_{i,t+1}\mid x_{i,\le t})}
{\displaystyle\sum_{i=1}^{N}\sum_{t=1}^{T_i-1}m_{i,t}}
\right),}
\label{eq:masked_ppl}
\end{equation}
where $m_{i,t}\in\{0,1\}$ identifies valid shifted target positions
in evaluation sample $i$. The same held-out samples, tokenization, and
masking are used for reference and quantized models.

Let \(S(r,\mathcal S)\) denote compressed storage and \(D(r,\mathcal S)\) a
quality-loss measure. We characterize the following conceptual storage--quality
operating-point objectives by evaluating configured residual depths and
projection scopes, rather than solving a constrained optimization problem:
\begin{align}
\min_{r,\mathcal S}\;&D(r,\mathcal S)
&&\mathrm{s.t.}\quad S(r,\mathcal S)\le B,\\
\min_{r,\mathcal S}\;&S(r,\mathcal S)
&&\mathrm{s.t.}\quad \PPL(r,\mathcal S)\le P_{\max}.
\end{align}
Thus \(r\) controls codec capacity and \(\mathcal S\) the type and fraction of
Transformer weights compressed.

\section{Proposed Method}
\label{sec:method}

\subsection{Overview and Activation-Aware Calibration}
The training-free pipeline selects attention, MLP, or attention+MLP
projections, collects calibration activations, applies optional
activation-aware equivalent scaling, and blockwise encodes selected weights
with the CDF-free residual-sign codec. No model parameters are optimized.

For input-channel activation statistic \(a_j\) and reference magnitude
\(a_{\mathrm{ref}}\), we define
\begin{equation}
s_j=
\left(\frac{a_j}{a_{\mathrm{ref}}}\right)^\beta,
\qquad
S=\operatorname{diag}(s_1,\ldots,s_{d_{\mathrm{in}}}),
\end{equation}
subject to configured bounds. The equivalent transformation is written as
\begin{align}
W' &= WS, \\
x' &= S^{-1}x, \\
W'x' &= Wx,
\end{align}
which changes the weight distribution presented to the codec without
changing the full-precision operation.



\subsection{CDF-Aware Quantization}
\label{sec:cdf_aware_codec}

When the cumulative distribution function (CDF) $F_X$ is known, a $b$-bit quantizer can map $X$ to a uniform variable $U=F_X(X)$ and assign
\begin{equation}
q(X)=\min\left\{2^b-1,
\left\lfloor 2^b F_X(X)\right\rfloor\right\}.
\end{equation}

Reconstruction uses the inverse CDF:

\begin{equation}
\widehat{X}=F_X^{-1}
\left(\frac{q(X)+1/2}{2^b}\right).
\end{equation}
This produces equal-probability bins but does not necessarily minimize mean-squared error. It also requires CDF knowledge or estimation. In contrast, our framework avoids explicit CDF estimation through orthogonal mixing, sign encoding, and progressive residual reconstruction.

\subsection{CDF-Free Residual-Sign Codec}
\label{sec:cdf_codec}

We now eliminate the need to know CDFs. For each length-$n$ residual vector $e$ in a row block, the codec computes its mean and centered RMS:
\begin{equation}
\mu=\frac{1}{n}\sum_{i=1}^{n}e_i,\qquad
\sigma=\sqrt{\frac{1}{n}\|e-\mu\mathbf1\|_2^2}.
\end{equation}
Using $q\in\{16,32\}$-bit stored statistics, let
$\widetilde\mu=Q_q(\mu)$ and
$\widetilde\sigma=Q_q(\max\{\sigma,\epsilon\})$,
where $Q_q$ denotes rounding to the specified format.
The residual is normalized, orthogonally mixed, and sign-encoded:
\begin{align}
z&=\frac{e-\widetilde\mu\mathbf1}
{\max\{\widetilde\sigma,\epsilon\}},\\
u&=Vz,\qquad VV^T=I_m,\\
s_i&=\begin{cases}
+1,&u_i\ge0,\\
-1,&u_i<0.
\end{cases}
\end{align}
Only the first $m_{\mathrm{eff}}\le m$ signs are stored;
the remaining transformed coordinates are zeroed during
reconstruction. By default, $m_{\mathrm{eff}}=m=n$.

\paragraph{Gaussian-CLT Motivation.}
Each mixed coordinate $u_i=\sum_jV_{ij}z_j$ may exhibit
approximately Gaussian behavior under suitable
weak-dependence and non-dominance conditions.
This motivates orthogonal mixing but neither guarantees
Gaussianity nor imposes a distributional assumption.
The transform is generated by QR orthonormalization of
a seeded Gaussian random matrix. Unlike QTIP
\cite{tseng2024qtip}, which employs trellis-coded
quantization, our codec directly encodes transformed
signs without Gaussian CDF or inverse-CDF evaluation.

For reconstruction, let $\bar s\in\mathbb R^m$ contain
the retained signs followed by zeros. Define
\begin{equation}
b=\widetilde\sigma V^T\bar s,\qquad
t=e-\widetilde\mu\mathbf1.
\end{equation}
A per-vector, per-stage least-squares gain is fitted
and stored at precision $q$:
\begin{align}
\gamma_{\mathrm{LS}}
&=\frac{t^Tb}{\max\{b^Tb,\epsilon\}},\\
\widetilde\gamma&=Q_q(\gamma_{\mathrm{LS}}),\\
\hat e&=\widetilde\mu\mathbf1+\widetilde\gamma b.
\end{align}
Nonfinite gains are replaced by the Gaussian-derived
fallback $\sqrt{2/\pi}$ before rounding.

For each row block $B$, the codec evaluates $K$
deterministic candidate transforms and selects
\begin{equation}
j^*=\arg\min_{1\le j\le K}
\frac{\|B-\widehat B(V_j)\|_F^2}{|B|},
\label{eq:transform_candidate_selection}
\end{equation}
where $\widehat B(V_j)$ includes progressive residual
refinement and any early termination. Only the selected
transform identifier and encoded signs and statistics
$(\widetilde\mu,\widetilde\sigma,\widetilde\gamma)$
are retained. No empirical CDF tables, learned
codebooks, or trellis paths are required.

\subsection{Successive Residual Refinement}
\label{sec:successive_refinement}
Successive residual-stage quantization progressively encodes reconstruction errors through sign-coded corrections, improving weight fidelity. Its progressive representation, and near-optimality at every level, enables inference-time selection of decoding depth, providing flexible storage--fidelity trade-offs without retraining or requantization. \\
For each row block $B$, initialize $E^{(0)}=B$ and $\widehat{B}^{(0)}=0$. At stage $k$, vectorwise encoding and decoding of $E^{(k-1)}$ yields a correction $\widehat{E}^{(k)}$ and
\begin{align}
\widehat{B}^{(k)}_{\mathrm{cand}}
&=\widehat{B}^{(k-1)}+\widehat{E}^{(k)},\\
E^{(k)}_{\mathrm{cand}}
&=B-\widehat{B}^{(k)}_{\mathrm{cand}}.
\end{align}
With $\mathrm{RMS}(A)=\sqrt{\|A\|_F^2/|A|}$, monotonic refinement accepts a stage only if
\begin{equation}
\mathrm{RMS}(E^{(k-1)})
-\mathrm{RMS}(E^{(k)}_{\mathrm{cand}})>\delta,
\end{equation}
where $\delta\geq0$ (default $0$). Accepted candidates become $\widehat{B}^{(k)}$ and $E^{(k)}$; otherwise, block encoding terminates. For $r_{\mathrm{enc}}\leq r$ accepted stages,
\begin{equation}
\widehat{B}^{(r_{\mathrm{enc}})}
=\sum_{k=1}^{r_{\mathrm{enc}}}\widehat{E}^{(k)},
\qquad
E^{(r_{\mathrm{enc}})}
=B-\widehat{B}^{(r_{\mathrm{enc}})}.
\end{equation}
Any accepted prefix is decodable, and acceptance is evaluated blockwise rather than vectorwise.

\paragraph{Near Optimality of Proposed Progressive CLT-VQ.}
With i.i.d. input weights, the orthogonally mixed, normalized residual coordinates are i.i.d.\ $u_i\sim\mathcal{N}(0,1)$. For constant-amplitude sign reconstruction $\widehat{u}_i=c\,\operatorname{sign}(u_i)$, the expected MSE is
\begin{equation}
D(c)=1-2c\sqrt{\frac{2}{\pi}}+c^2.
\end{equation}
The optimal amplitude is $c^*=\sqrt{2/\pi}$, yielding $D(c^*)=1-2/\pi\approx0.3634$, a $63.7\%$ reduction in expected squared error within this reconstruction family.

Assuming i.i.d. input and CLT mixing, all successive stages' residuals are conveniently also i.i.d. Gaussian, hence, successive refinement compounds distortion like so:
\begin{equation}
\frac{D_k}{n}=\left(1-\frac{2}{\pi n}\right)D_{k-1},
\;
\frac{D_L}{n}=\left(1-\frac{2}{\pi n}\right)^L D_0.
\end{equation}
As $n \gg 1$, $(1 - 1/n)^n \to e^{-1}$ giving $\frac{D_L}{n}\approx D_0 2^{-0.9185 \frac{L}{n}}$.

On the other hand, for a memoryless (i.i.d.) Gaussian source, Shannon's rate--distortion function is worst, giving the information theoretic distortion lower-bound of:
\begin{equation}
\frac{D_{\mathrm{RD}}(R)}{n} \ge D_0 2^{-2R},
\end{equation}
where rate $R = \tfrac{L}{n}$ denotes bits per coordinate. Thus, the predicted decay exponent $0.9185$ is approximately $45\%$ of the optimal Gaussian rate--distortion exponent, while still preserving exponential improvement with $L$.

Our implementation uses least-squares reconstruction gains and blockwise monotonic acceptance to ensure non-increasing realized block distortion. However, if the input weights are not i.i.d., residuals are not automatically guaranteed to be Gaussian across stages -- but as CLT is applicable under relatively mild conditions, this can be remedied by normalization and whitening of input weights, while stored statistics metadata may increase the effective rate.

\subsection{Blockwise Execution and Storage}
\label{sec:parallel_codec}

Selected matrices are partitioned into independent row blocks with associated range, projection type, residual stream, statistics, and transform metadata. For $P$ codec devices, blocks are assigned as
\begin{equation}
B_i \mapsto G_{\,1+((i-1)\bmod P)},
\end{equation}
enabling multi-GPU quantization. Compressed blocks are reconstructed on demand with optional caching, while both models can be evaluated on a single accelerator.

Checkpoints may store compressed projections and metadata as overlays, loading untargeted parameters from the pretrained model. Thus, selected weight payload, serialized overlay size, and complete deployment storage are reported separately, since overlay size does not represent a self-contained compressed model.

\label{sec:quant_algorithm}
\begin{algorithm}[t]
\caption{CDF-Free Progressive Residual-Sign Quantization}
\label{alg:cdf_quantization}
\footnotesize
\begin{algorithmic}[1]

\REQUIRE Model $f$, scope $\mathcal{S}$, calibration data, stages $r$, candidates $K$, statistic bits $q$, threshold $\delta$

\STATE Select projections and collect activation statistics
\STATE Compute scales and form $W'=WS$

\FOR{each selected matrix $W'$}
    \STATE Partition $W'$ into row blocks $B$

    \FOR{each block $B$}
        \STATE Generate transforms $\{V_j\}_{j=1}^{K}$

        \FOR{$j=1,\ldots,K$}
            \STATE $E\gets B$, $\widehat{B}_j\gets 0$

            \FOR{$k=1,\ldots,r$}
                \STATE $\widehat{E}\gets
                \mathrm{SignEncodeDecode}(E,V_j,q)$

                \STATE $E_{\mathrm{new}}\gets E-\widehat{E}$

                \IF{$\mathrm{monotonic}$ and
                $\mathrm{RMS}(E)-
                \mathrm{RMS}(E_{\mathrm{new}})\leq\delta$}
                    \STATE \textbf{break}
                \ENDIF

                \STATE $\widehat{B}_j\gets
                \widehat{B}_j+\widehat{E}$
                \STATE $E\gets E_{\mathrm{new}}$
            \ENDFOR

            \STATE $D_j\gets
            \|B-\widehat{B}_j\|_F^2/|B|$
        \ENDFOR

        \STATE $j^*\gets\arg\min_j D_j$
        \STATE Re-encode $B$ using $V_{j^*}$
        \STATE Pack accepted stages and metadata

    \ENDFOR
\ENDFOR

\STATE Save compressed overlay and streaming modules

\end{algorithmic}
\end{algorithm}

\FloatBarrier

\section{Experimental Methodology}
\label{sec:experimental_methodology}

We evaluate GPT-2 Large (\texttt{gpt2-large}) and Llama-3-8B
(\texttt{meta-llama/Meta-Llama-3-8B}) under attention, MLP, and
attention+MLP quantization, excluding embeddings and the language-model head.
Each scope uses \(r\in\{1,2,3,5\}\), yielding
\(2\times3\times4=24\) primary conditions. 

Calibration, test, and held-out subsets are constructed from WikiText-2 raw
(\texttt{wikitext-2-raw-v1}). Original splits are combined, normalized,
filtered, deduplicated, and deterministically shuffled (seed 42); the first
512 samples form the calibration pool, the next 256 the test set, and the
final 256 the held-out set. Calibration only estimates activation statistics;
runs use at most 32 calibration samples, and manuscript metrics use the
held-out set. The supplied Llama-3-8B evaluation JSON identifies
\texttt{data/evaluation.txt}, a limit of 256 evaluation samples, a maximum
sequence length of 256, and 36,153 evaluated tokens. The JSON alone does
not independently verify the provenance of the evaluation file.

\begin{table}[!tbp]
\centering
\caption{Primary quantization configuration.}
\label{tab:codec_config}
\small
\setlength{\tabcolsep}{4pt}
\begin{tabular}{ll}
\toprule
\textbf{Parameter} & \textbf{Setting} \\
\midrule
Vector/projection dimensions \((n,m)\) & \((256,256)\) \\
Residual stages \(r\) & \(1,2,3,5\) \\
Statistics precision & 32 bits \\
Block batch size & 4 \\
Activation-aware scaling & \(\beta=0.5\) \\
Transform candidates \(K\) & 4 \\
Calibration samples / seq. length & 32 / 256 \\
Progressive / entropy coding & Enabled / Enabled \\
Random seed & 42 \\
\bottomrule
\end{tabular}
\end{table}

The full-precision model is the reference. For Llama-3-8B, the reported
checkpoint sizes are measured serialized overlay sizes; they do not include
all unchanged full-precision parameters. Llama-3-8B runtime measurements
use model-sharded execution and should not be interpreted as optimized
compressed-kernel performance.\par
Parameter metrics include MSE,
RMSE, MAE, relative error, cosine similarity, PSNR, and SNR; functional
metrics include logit MSE/cosine similarity and next-token perplexity under
matched tokenization and samples. Storage metrics include selected-weight
payload, compression ratio, bits per parameter, and serialized checkpoint
size. Because checkpoints can be overlays, their size is distinguished from
self-contained deployment storage. We also record quantization/evaluation
runtime and GPU memory where available.

Evaluation of Llama-3-8B uses model sharding across four GPUs as recorded in the evaluation metadata; the GPT-2 Large execution setup is unchanged.

\paragraph{Separate baseline-comparison protocol.} The additional joint-scope experiment (Appendix~B) uses a separate FP16 evaluation with 4,096 input tokens (4,095 scored tokens), selected-weight estimated payload BPW, and reconstructed dense weights. Its rate and perplexity measurements are not directly comparable to the scope-specific streaming-overlay experiments.

\section{Results}
\label{sec:results}

We evaluate GPT-2 Large and Llama-3-8B at residual depths $r\in\{1,2,3,5\}$ for attention-only, MLP-only, and joint attention+MLP quantization. Metrics include reconstruction and logit errors, token agreement, perplexity, serialized overlay size, and streaming runtime.

\subsection{GPT-2 Large}
\label{sec:results_gpt2}

Full-precision GPT-2 Large yields perplexity
\(29.89\) on held-out data. All three quantization scopes
use the same evaluation protocol, enabling direct comparison across
scope and residual depth.




\begin{figure*}[!t]
    \centering
    \includegraphics[
        width=0.85\textwidth,
        keepaspectratio
    ]{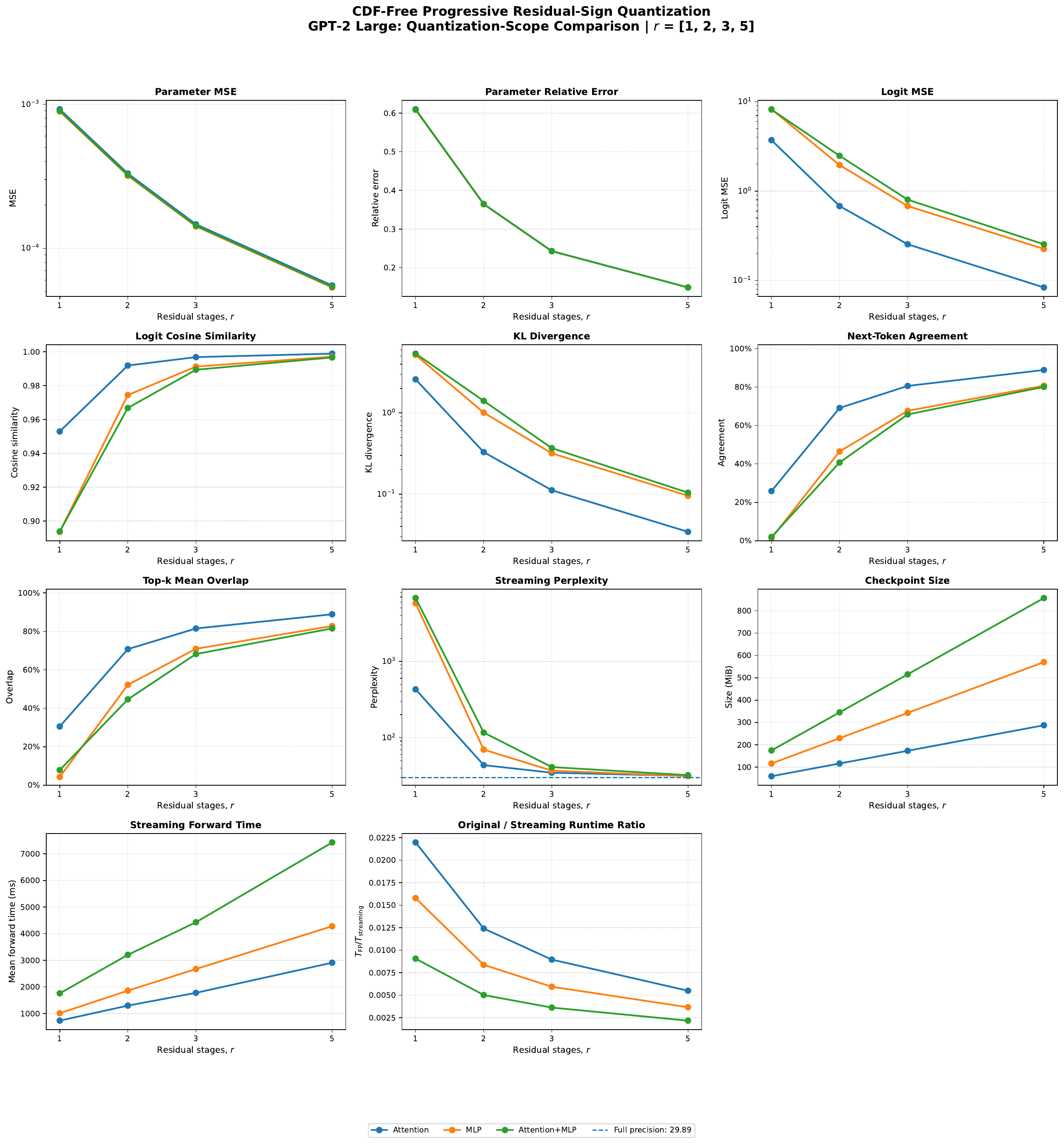}

    \caption{
        Quantization-scope comparison for GPT-2 Large
        across residual depths $r\in\{1,2,3,5\}$.
    }
    \label{fig:gpt2_scope_comparison}
\end{figure*}

\begin{table*}[!tp]
\centering
\caption{GPT-2 Large comparison across all three quantization scopes.
The full-precision perplexity is \(29.89\). Runtime ratio is
\(T_{\mathrm{FP}}/T_{\mathrm{streaming}}\).}
\label{tab:gpt2_scope_comparison}
\resizebox{\textwidth}{!}{
\begin{tabular}{llrrrrrrrrrrr}
\toprule
Scope & \(r\) &
Param. MSE &
Rel. Err. &
Logit MSE &
Logit Cos. &
KL &
Next (\%) &
Top-\(k\) (\%) &
PPL &
Size (MiB) &
Forward (ms) &
Runtime Ratio\\
\midrule

Attention & 1 & \(9.225\times10^{-4}\) & 0.6100 & 3.7176 & 0.9530 & 2.5918 & 25.85 & 30.53 & 429.36 & 58.83 & 731.19 & 0.0220\\
Attention & 2 & \(3.303\times10^{-4}\) & 0.3650 & 0.6805 & 0.9919 & 0.3289 & 69.11 & 70.77 & 43.58 & 115.97 & 1295.36 & 0.0124\\
Attention & 3 & \(1.467\times10^{-4}\) & 0.2432 & 0.2542 & 0.9969 & 0.1114 & 80.56 & 81.52 & 34.64 & 172.95 & 1776.10 & 0.0090\\
Attention & 5 & \(5.519\times10^{-5}\) & 0.1492 & 0.0836 & 0.9990 & 0.0342 & 88.86 & 88.91 & 31.44 & 287.32 & 2909.18 & 0.0055\\
\midrule

MLP & 1 & \(8.912\times10^{-4}\) & 0.6088 & 8.2450 & 0.8936 & 5.1919 & 1.35 & 4.23 & 5765.20 & 115.80 & 1010.37 & 0.0158\\
MLP & 2 & \(3.192\times10^{-4}\) & 0.3644 & 1.9571 & 0.9745 & 1.0087 & 46.47 & 52.21 & 69.61 & 229.37 & 1858.88 & 0.0084\\
MLP & 3 & \(1.418\times10^{-4}\) & 0.2429 & 0.6814 & 0.9913 & 0.3180 & 67.61 & 70.96 & 36.94 & 342.71 & 2671.84 & 0.0059\\
MLP & 5 & \(5.348\times10^{-5}\) & 0.1491 & 0.2256 & 0.9972 & 0.0952 & 80.68 & 82.81 & 31.15 & 570.18 & 4281.04 & 0.0037\\
\midrule

Attention+MLP & 1 & \(9.016\times10^{-4}\) & 0.6092 & 8.1806 & 0.8938 & 5.3807 & 1.97 & 7.86 & 6768.31 & 174.38 & 1757.09 & 0.0091\\
Attention+MLP & 2 & \(3.229\times10^{-4}\) & 0.3646 & 2.4858 & 0.9668 & 1.4073 & 40.73 & 44.64 & 116.48 & 345.09 & 3204.37 & 0.0050\\
Attention+MLP & 3 & \(1.434\times10^{-4}\) & 0.2430 & 0.8028 & 0.9894 & 0.3682 & 65.74 & 68.23 & 41.09 & 515.42 & 4429.88 & 0.0036\\
Attention+MLP & 5 & \(5.405\times10^{-5}\) & 0.1492 & 0.2542 & 0.9967 & 0.1041 & 80.05 & 81.55 & 32.16 & 857.25 & 7430.52 & 0.0022\\

\bottomrule
\end{tabular}}
\end{table*}

\subsubsection{Comparison Across Quantization Scopes}
\label{sec:gpt2_scope_analysis}

Figure~\ref{fig:gpt2_scope_comparison} and
Table~\ref{tab:gpt2_scope_comparison} show a consistent progressive
recovery trend across all three scopes. Parameter MSE and relative error
decrease similarly with \(r\). At \(r=5\),
parameter MSE is approximately \(5.3\text{--}5.5\times10^{-5}\) and relative
error is approximately \(0.149\) across all scopes. The main differences
therefore appear in model-level fidelity rather than in the local
weight-reconstruction trend.

Attention-only quantization is the least disruptive scope at low and
moderate residual depth. At \(r=1\), its perplexity is \(429.36\),
compared with \(5765.20\) for MLP-only and \(6768.31\) for joint
attention+MLP quantization. The gap narrows rapidly with additional
residual stages: at \(r=3\), perplexity is \(34.64\), \(36.94\), and
\(41.09\) for attention, MLP, and attention+MLP, respectively. At
\(r=5\), all three configurations are close to the full-precision baseline,
with perplexities of \(31.44\), \(31.15\), and \(32.16\), respectively.

Output fidelity also converges. At \(r=5\),
attention-only quantization gives the strongest overall logit fidelity,
with logit MSE \(0.0836\), cosine similarity \(0.9990\), KL divergence
\(0.0342\), next-token agreement \(88.86\%\), and top-\(k\) overlap
\(88.91\%\). MLP-only and joint quantization also recover high fidelity,
with next-token agreement of \(80.68\%\) and \(80.05\%\), respectively.
MLP-only quantization obtains the lowest perplexity at this retained maximum
stage count, \(31.15\), while attention-only quantization retains stronger
token-level and logit-level agreement.

Storage and runtime scale primarily with the count of quantized
parameters. Attention-only quantizes approximately \(236\) million
parameters, MLP-only approximately \(472\) million, and the joint scope
approximately \(708\) million. Consequently, at \(r=5\), checkpoint size
increases from \(287.32\) MiB for attention to \(570.18\) MiB for MLP
and \(857.25\) MiB for attention+MLP. Streaming forward time follows the
same ordering: approximately \(2.91\) s, \(4.28\) s, and \(7.43\) s.
These results show that residual depth controls fidelity within a scope,
while scope size determines the dominant storage and runtime cost.

Overall, \(r=5\), the largest residual depth retained in this study,
provides near-baseline perplexity for all three scopes. The results therefore
support a clear rate--distortion interpretation: increasing residual depth
through the evaluated range recovers model quality, whereas quantizing a
larger subsystem increases the cost required to reach a comparable fidelity
level.


\subsection{Llama-3-8B}
\label{sec:results_llama}

The full-precision Llama-3-8B model (\texttt{meta-llama/Meta-Llama-3-8B})
obtains perplexity \(11.07\) on the supplied evaluation set (36,153
scored tokens). We evaluate attention-only, MLP-only, and joint
attention+MLP quantization at \(r\in\{1,2,3,5\}\). The scopes contain
1,342,177,280, 5,637,144,576, and 6,979,321,856 selected
parameters, respectively. These are scope-specific quantization
experiments rather than whole-model self-contained compression rates.



\begin{figure*}[!t]
    \centering
    \includegraphics[
        width=0.90\textwidth,
        keepaspectratio
    ]{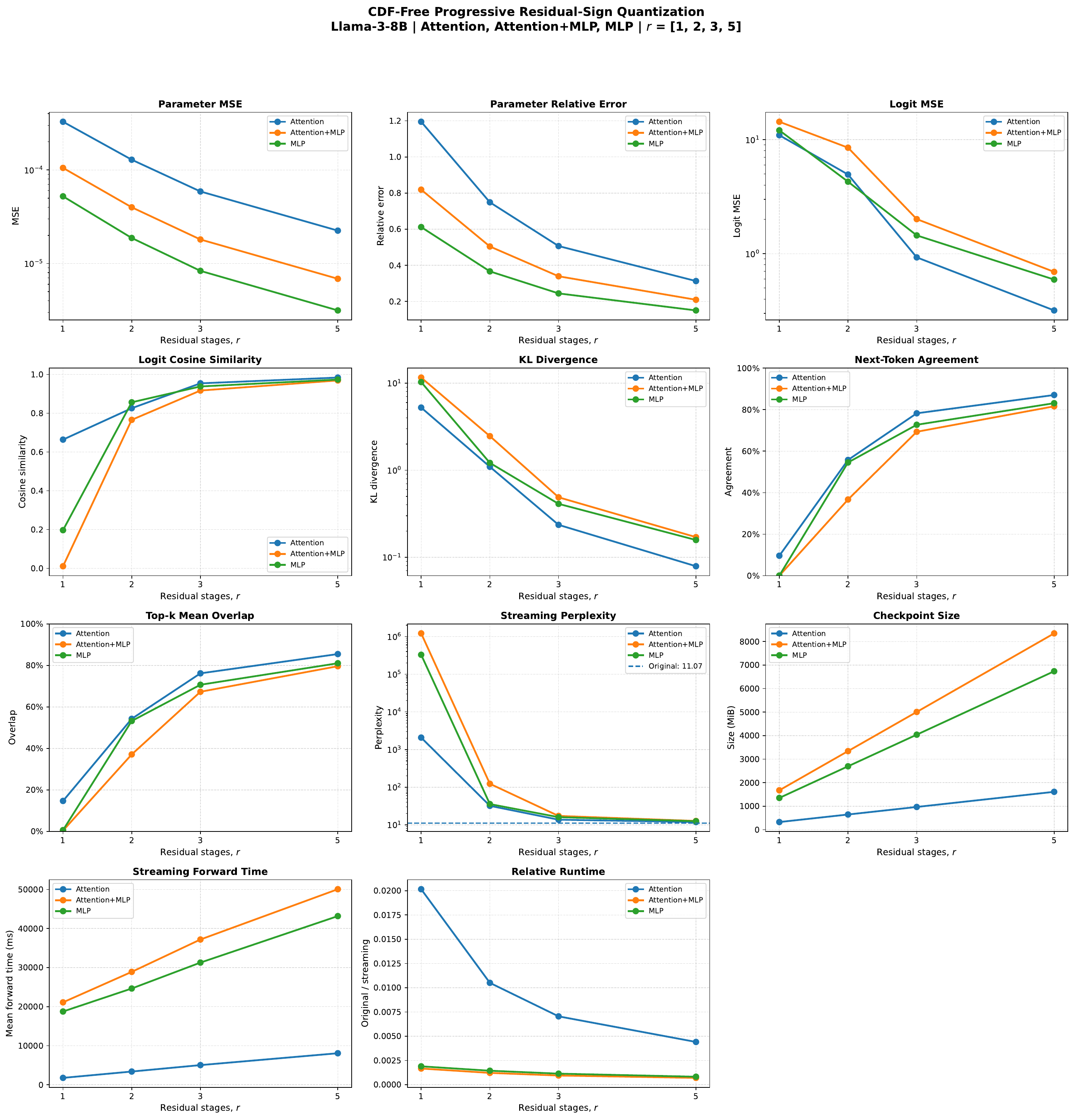}

    \caption{
        Quantization-scope comparison for Llama-3-8B
        across residual depths $r\in\{1,2,3,5\}$.
    }
    \label{fig:llama_scope_comparison}
\end{figure*}

\begin{table*}[!tp]
\centering
\caption{Llama-3-8B scope results (FP PPL: 11.07; runtime ratio: $T_{\mathrm{FP}}/T_{\mathrm{streaming}}$).}
\label{tab:llama_scope_comparison}
\resizebox{\textwidth}{!}{
\begin{tabular}{llrrrrrrrrrrr}
\toprule
Scope & \(r\) & Param. MSE & Rel. Err. & Logit MSE & Logit Cos. & KL & Next (\%) & Top-\(k\) (\%) & PPL & Size (MiB) & Forward (ms) & Runtime Ratio\\
\midrule
Attention & 1 & \(3.278e-04\) & 1.1949 & 10.9461 & 0.6640 & 5.2465 & 9.62 & 14.72 & 2076.92 & 324.62 & 1762.50 & 0.0202\\
Attention & 2 & \(1.288e-04\) & 0.7490 & 4.9474 & 0.8259 & 1.0956 & 55.77 & 54.29 & 32.09 & 645.72 & 3382.69 & 0.0105\\
Attention & 3 & \(5.892e-05\) & 0.5066 & 0.9293 & 0.9543 & 0.2358 & 78.24 & 76.20 & 13.75 & 966.73 & 5030.83 & 0.0070\\
Attention & 5 & \(2.247e-05\) & 0.3129 & 0.3185 & 0.9840 & 0.0789 & 87.06 & 85.49 & 11.90 & 1610.37 & 8067.03 & 0.0044\\
\midrule
MLP & 1 & \(5.228e-05\) & 0.6118 & 12.0770 & 0.1970 & 10.3045 & 0.05 & 0.66 & 327947.62 & 1350.58 & 18748.79 & 0.0019\\
MLP & 2 & \(1.875e-05\) & 0.3664 & 4.2837 & 0.8565 & 1.2143 & 54.55 & 53.25 & 35.70 & 2696.00 & 24639.92 & 0.0014\\
MLP & 3 & \(8.345e-06\) & 0.2444 & 1.4481 & 0.9382 & 0.4110 & 72.71 & 70.70 & 16.04 & 4040.53 & 31273.26 & 0.0011\\
MLP & 5 & \(3.152e-06\) & 0.1502 & 0.5938 & 0.9738 & 0.1579 & 83.11 & 81.06 & 12.55 & 6737.66 & 43173.52 & 0.0008\\
\midrule
Attention+MLP & 1 & \(1.053e-04\) & 0.8189 & 14.3613 & 0.0109 & 11.6030 & 0.00 & 0.42 & 1211639.22 & 1674.99 & 21097.99 & 0.0017\\
Attention+MLP & 2 & \(3.991e-05\) & 0.5043 & 8.5083 & 0.7659 & 2.4650 & 36.70 & 37.13 & 123.44 & 3341.50 & 28899.10 & 0.0012\\
Attention+MLP & 3 & \(1.807e-05\) & 0.3393 & 2.0122 & 0.9171 & 0.4889 & 69.33 & 67.32 & 17.29 & 5007.07 & 37183.52 & 0.0009\\
Attention+MLP & 5 & \(6.867e-06\) & 0.2092 & 0.6937 & 0.9692 & 0.1704 & 81.61 & 79.64 & 12.73 & 8347.85 & 50059.68 & 0.0007\\
\bottomrule
\end{tabular}}
\end{table*}

\subsubsection{Comparison Across Quantization Scopes}
\label{sec:llama_scope_analysis}
Figure~\ref{fig:llama_scope_comparison} and
Table~\ref{tab:llama_scope_comparison} demonstrate progressive recovery
at every scope. At \(r=1\), perplexity is \(2076.92\) for attention,
\(327947.62\) for MLP, and \(1.21\times10^{6}\) for joint
quantization. Thus, a single residual stage does not preserve usable
language-model fidelity under the present settings, particularly when
MLP weights are quantized. Next-token agreement is \(9.62\%\),
\(0.05\%\), and \(0.00\%\), respectively.

At \(r=2\), perplexity decreases to \(32.09\), \(35.70\), and
\(123.44\) for attention, MLP, and joint scopes. At \(r=3\), it
reaches \(13.75\), \(16.04\), and \(17.29\). With five residual
stages, the corresponding perplexities are \(11.90\), \(12.55\), and
\(12.73\), compared with \(11.07\) for full precision. The relative
increases at \(r=5\) are approximately \(7.5\%\), \(13.3\%\), and
\(15.0\%\), respectively.

Output fidelity improves alongside perplexity. At \(r=5\), the
attention-only setting attains logit cosine similarity \(0.9840\),
KL divergence \(0.0789\), next-token agreement \(87.06\%\), and
top-\(k\) overlap \(85.49\%\). MLP-only obtains cosine similarity
\(0.9738\), agreement \(83.11\%\), and top-\(k\) overlap
\(81.06\%\). Joint quantization obtains cosine similarity \(0.9692\),
agreement \(81.61\%\), and top-\(k\) overlap \(79.64\%\).
Parameter MSE decreases with stage depth within each scope: from
\(3.278\times10^{-4}\) to \(2.247\times10^{-5}\) for attention,
\(5.228\times10^{-5}\) to \(3.152\times10^{-6}\) for MLP, and
\(1.053\times10^{-4}\) to \(6.867\times10^{-6}\) for joint
quantization.

At \(r=5\), serialized overlay checkpoints occupy \(1610.37\) MiB,
\(6737.66\) MiB, and \(8347.85\) MiB for attention, MLP, and joint
scopes, respectively. Mean streaming forward times are \(8.07\) s,
\(43.17\) s, and \(50.06\) s, respectively. The streaming path is
substantially slower than the full-precision reference (approximately
35 ms per forward pass); these measurements do not demonstrate
accelerated inference. The checkpoints also retain dependencies on the
unchanged base-model weights and should not be interpreted as
self-contained deployment artifacts.

\subsection{Cross-Model Analysis}
\label{sec:cross_model_results}

GPT-2 Large and Llama-3-8B exhibit the same broad trend: increasing
residual depth reduces parameter distortion and improves language-model
fidelity for attention-only, MLP-only, and joint quantization. However,
their baseline perplexities and quantized parameter counts differ, so
absolute perplexities and raw checkpoint sizes should not be interpreted
as matching comparisons. GPT-2 Large has a full-precision perplexity
of \(29.89\), and Llama-3-8B has \(11.07\).

At \(r=5\), GPT-2 Large perplexity is \(31.44\), \(31.15\), and
\(32.16\) for attention, MLP, and joint scopes, corresponding to
approximately \(5.2\%\), \(4.2\%\), and \(7.6\%\) above baseline.
For Llama-3-8B the corresponding perplexities are \(11.90\),
\(12.55\), and \(12.73\), or \(7.5\%\), \(13.3\%\), and \(15.0\%\)
above baseline. Attention-only quantization is the least disruptive
scope at \(r=1\) for both models, but differences among scopes remain
visible at \(r=5\), particularly for Llama-3-8B.

The larger Llama-3-8B quantized scopes also incur substantially larger
serialized overlays and longer streaming reconstruction times. These
figures characterize the measured prototype, not a normalized
compression advantage or hardware speedup. The results support residual
depth as a useful fidelity control while underscoring the need for
scope-aware and implementation-aware evaluation.

\subsection{Orthogonal-Mixing Comparison}
\label{sec:orthogonal_benchmark_results}
\begin{figure}[!tbp]
\centering
\includegraphics[width=\columnwidth,height=0.29\textheight,keepaspectratio]{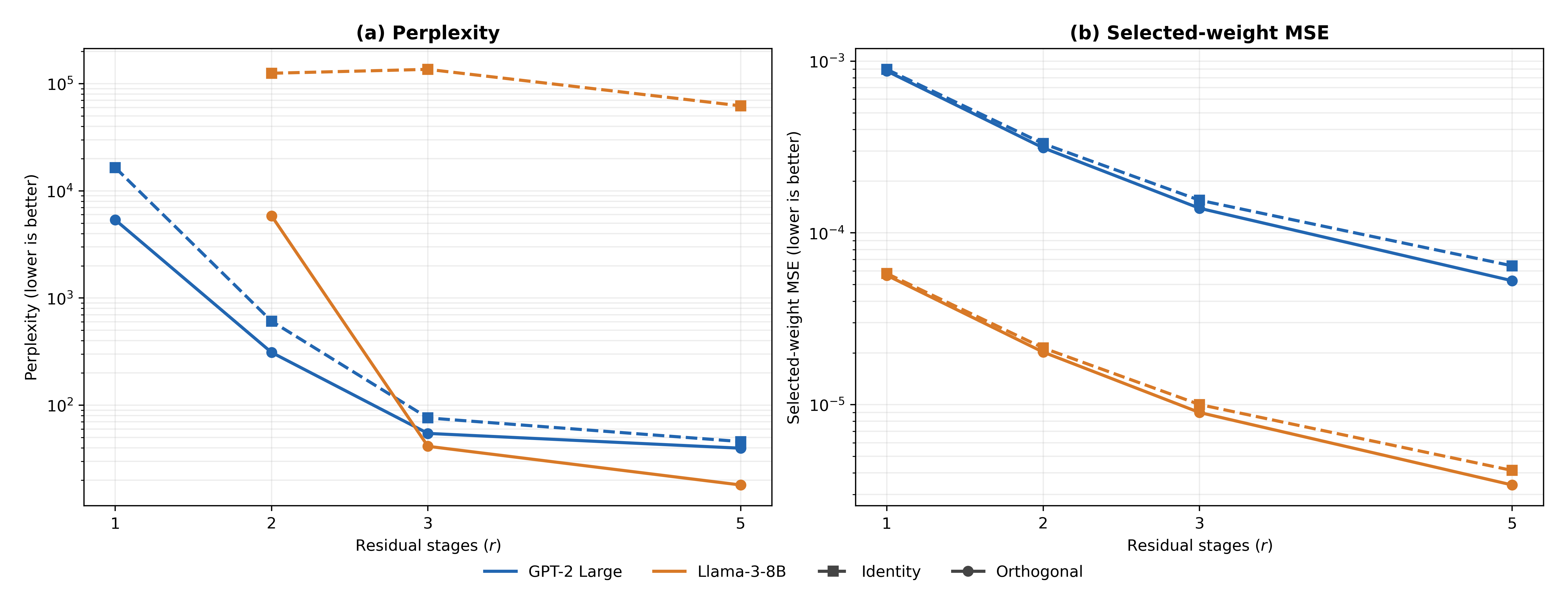}
\caption{Orthogonal mixing and model fidelity across GPT-2 Large and Llama-3-8B.}
\label{fig:orthogonal_identity_combined}
\end{figure}

A separate joint-scope benchmark evaluates FP16 reconstructed dense weights on 4,095 scored tokens; its results are not directly comparable to the scope-specific streaming runs. Figure~\ref{fig:orthogonal_identity_combined} shows that orthogonal mixing lowers GPT-2 Large perplexity and selected-weight MSE at all tested depths. On Llama-3-8B, MSE improves at every depth and perplexity at $r=2,3,5$; the omitted $r=1$ perplexity is worse with orthogonal mixing ($2.42\times10^6$ versus $9.03\times10^5$). Estimated payloads are nearly matched between the two sign variants.

\paragraph{Scalar and vector baselines.}
The orthogonal sign codec does not dominate conventional quantizers in perplexity per selected-weight payload. At $r=5$, GPT-2 Large sign coding yields PPL $39.73$ at $6.618$ estimated BPW, whereas scalar 4-bit coding at $r=1$ yields $37.59$ at $3.658$ BPW and residual VQ ($K=256$, $r=5$) yields $36.98$ at $4.995$ BPW. For Llama-3-8B, sign coding gives $18.02$ at $6.611$ BPW, compared with scalar 4-bit's $16.05$ at $3.632$ BPW; the tested VQ ($K=256$, $r=5$) instead gives PPL $8449$ at $4.948$ BPW. Thus, the benchmark supports orthogonal mixing and progressive sign refinement, not a universally better rate--distortion frontier. The full comparison and measurement caveats appear in Appendix~B.


\section{Ablation Studies}
\label{sec:ablation}
Both models use joint attention+MLP quantization at $r=5$ with $n=m=256$, 32-bit statistics, 128 block rows, block batch size 4, $K=4$, activation-aware scaling (AAS), 32 calibration samples, monotonic stage acceptance, and seed 42 unless otherwise stated. Detailed measurements appear in Appendix~A.

For GPT-2 Large, AAS reduces logit MSE from $0.3001$ to $0.2543$ and perplexity increase from $9.79\%$ to $7.63\%$, despite higher parameter MSE. Four transform candidates outperform one; calibration-size and block-row effects are modest. Using 16-bit rather than 32-bit statistics increases compression from $4.77\times$ to $5.53\times$ with negligible fidelity change. Vector dimension controls the observed storage--fidelity trade-off.

For Llama-3-8B, AAS reduces logit MSE from $1.0983$ to $0.6920$ and PPL from $13.9445$ to $12.5854$, although parameter MSE rises from $3.379\times10^{-6}$ to $6.867\times10^{-6}$. Sixteen-bit statistics reduce the overlay from $8347.85$ to $7220.71$ MiB with comparable PPL. Increasing $K$ from 1 to 4 reduces PPL from $12.6102$ to $12.5854$ but increases encoding time from about $100.3$ to $175.6$ s. Other tested settings have smaller or model-dependent effects. No Llama-3-8B vector-dimension or early-stop-rate ablation is available. These results favor evaluating functional fidelity alongside weight error.

\section{Discussion}
\label{sec:discussion}

\paragraph{Coding, depth, and scope.}
Our proposed framework combines distortion-selected orthogonal mixing, analytical sign reconstruction, and progressive residual refinement without requiring learned codebooks or CDF estimation. Increasing $r$ trades storage for fidelity: from $r=1$ to $r=5$, joint-scope PPL drops from $6768.31$ to $32.16$ (GPT-2 Large) and $1.21\times10^6$ to $12.73$ (Llama-3-8B), versus full-precision $29.89$ and $11.07$. Attention-only coding is less disruptive at $r=1$; all scopes approach baseline at $r=5$ (Section~6).

\vspace*{-2ex}
\paragraph{Weight error and output fidelity.}
Weight MSE does not reliably predict functional quality. At $r=5$, Llama-3-8B attention-only and joint coding yield MSEs of $2.247\times10^{-5}$ and $6.867\times10^{-6}$ but PPLs of $11.90$ and $12.73$, respectively; their MSEs cover different parameter sets. GPT-2 Large ablations similarly show AAS benefits despite higher weight error. On Llama-3-8B, AAS raises weight MSE from $3.379\times10^{-6}$ to $6.867\times10^{-6}$ but lowers logit MSE from $1.0983$ to $0.6920$ and PPL from $13.9445$ to $12.5854$.

\vspace*{-2ex}
\paragraph{Calibration and operating points.}
Both models improve with depth but differ in low-depth sensitivity. Llama-3-8B uses AAS $\beta=0.5$ and four transform candidates. One-factor ablations cover AAS, candidate and calibration counts, statistics precision, block rows, and batch size, but not interactions, vector dimensions, or seed generalization. Lower $r$ and larger vectors favor storage; higher $r$ and smaller vectors favor fidelity. Across both models, $r=3$ restores substantial quality and $r=5$ approaches baseline; 16-bit statistics further reduce storage. Increasing depth or scope raises storage and reconstruction time. Streaming remains slower than full precision, while overlay size excludes self-contained deployment costs. Cross-scope rates therefore require selected-weight BPW or compression ratios.

\vspace*{-2ex}
\paragraph{Baselines and Gaussian motivation.}
Appendix~B compares identity/orthogonal sign coding, scalar quantization, and residual VQ at estimated selected-weight rates. Mixing often improves sign coding, although scalar quantization or VQ can achieve lower PPL at lower BPW; codebook-free coding guarantees neither inference speedup nor universal rate--distortion superiority. Finally, the CLT motivates mixing without assuming Gaussianity: reconstruction uses least-squares-fitted gains, not Gaussian CDF evaluation.





\section{Limitations}
\label{sec:limitations}

AAS relies on limited calibration data, with domain-shift robustness untested. Global residual depth ignores layer-specific sensitivity. Quantization scopes differ in parameter count, making overlay sizes incomparable without normalization and distinct from self-contained deployment storage. Streaming weight reconstruction is slower than full precision, while multi-GPU scaling depends on hardware and transfer overhead.

Evaluation is limited to GPT-2 Large and Llama-3-8B, primarily using perplexity and short-horizon fidelity; downstream, long-context, instruction-following, and generative robustness remain untested. Although Llama-3-8B includes scope/depth sweeps and component ablations, vector-dimension, early-stop, repeated-seed, and matched-rate baseline analyses are absent. One-factor ablations omit hyperparameter interactions, and broad hardware-matched PTQ comparisons remain unavailable. Finally, Gaussian-CLT behavior is a motivation rather than a verified result; orthogonality alone guarantees neither Gaussianity nor independence.

\section{Conclusion and Future Work}
\label{sec:conclusion}

We introduced training-free, CDF-free, Gaussian-CLT-motivated residual-sign quantization with orthogonal mixing and compact stage statistics.
Progressive reconstruction follows
\[
\boxed{\hat x^{(r)}=\sum_{k=1}^{r}\hat e^{(k)}},
\]
where residual depth $r$ controls the rate--distortion trade-off without learned codebooks, empirical CDFs, or trellis optimization.

Progressive refinement improves fidelity across attention-only, MLP-only, and joint quantization. At $r=5$, their respective perplexities are $31.44$, $31.15$, and $32.16$ for GPT-2 Large (full precision: $29.89$), and $11.90$, $12.55$, and $12.73$ for Llama-3-8B (full precision: $11.07$), with scope-dependent fidelity gaps.

Llama-3-8B ablations (Section~\ref{sec:ablation}; Appendix~A) show that activation-aware scaling improves perplexity from $13.9445$ to $12.5854$ and next-token agreement from $75.68\%$ to $81.75\%$, despite increased parameter MSE. Reducing statistics precision from 32 to 16 bits shrinks checkpoints from $8347.85$ to $7220.71$ MiB, with perplexities of $12.5854$ and $12.5803$, respectively. Four versus one transform candidate modestly improves MSE and perplexity at higher encoding cost, while calibration size and block-row/batch sizes minimally affect perplexity within tested ranges. These ablations exclude orthogonal-versus-identity comparisons and adaptive early stopping.

A separate matched-configuration joint-scope benchmark (Appendix~B) shows that, at $r=5$, orthogonal mixing reduces perplexity from $45.81$ to $39.73$ on GPT-2 Large and from approximately $6.22\times10^4$ to $18.02$ on Llama-3-8B, relative to identity sign coding at similar estimated selected-weight payloads. However, scalar quantization achieves lower perplexity at some higher payload rates, precluding claims of universal rate--distortion superiority.

Streaming reconstruction preserves compactness and fidelity but incurs substantial decode-and-forward overhead. Future work includes sensitivity- or residual-energy-guided layer- and projection-specific stage allocation,
\[
\boxed{
\min_{\{r_\ell\}}D_{\mathrm{functional}}
\quad\mathrm{s.t.}\quad
\sum_{\ell=1}^{L}S_\ell(r_\ell)\le B,
}
\]
fused decoding/matrix-multiplication kernels, improved transform selection, self-contained compressed checkpoints, broader downstream and long-context evaluations, additional model families, and matched post-training quantization baselines.

% ...
\FloatBarrier
\bibliographystyle{mlsys2025}
\begingroup
\small
\setlength{\emergencystretch}{4em}

\endgroup
\clearpage
\appendix
\section{Detailed Ablation Studies}
\label{app:ablation}

This appendix provides the complete component-level ablations summarized in Section~7. The main GPT-2 Large and Llama-3-8B results, including their scope-comparison tables, figures, and detailed interpretations, are presented in Section~6; they are not repeated here.

Unless stated otherwise, both models use joint attention+MLP quantization at
\(r=5\), \(n=m=256\), 32-bit statistics, 128 block rows, block batch size 4,
\(K=4\), activation-aware scaling (AAS), 32 calibration samples, monotonic
stage acceptance, and seed 42. Only the specified component varies.

For GPT-2 Large (Table~\ref{tab:ablation}), AAS improves functional fidelity
despite higher parameter MSE: logit MSE falls \(0.3001\to0.2543\)
and \(\Delta\PPL\) \(9.79\%\to7.63\%\). \(K=4\) improves over \(K=1\);
calibration size has modest effect. Using 16-bit rather than 32-bit statistics
raises compression \(4.77\times\to5.53\times\) with negligible quality
change. Block-row changes are small, while vector dimension exposes the
expected fidelity--compression trade-off.
\par\medskip
\noindent\begin{minipage}{\columnwidth}
\centering
\captionof{table}{GPT-2 Large component ablations ($r=5$, joint scope).}
\label{tab:ablation}
\resizebox{\columnwidth}{!}{%
\begingroup
\scriptsize
\setlength{\tabcolsep}{2.5pt}
\begin{tabular}{llrrrrr}
\hline
\textbf{Ablation} & \textbf{Setting} &
\textbf{Param. MSE}$\downarrow$ &
\textbf{Logit MSE}$\downarrow$ &
\textbf{$\Delta$PPL (\%)}$\downarrow$ &
\textbf{BPW}$\downarrow$ &
\textbf{Comp.}$\uparrow$ \\
\hline
AAS
 & Off & $5.204{\times}10^{-5}$ & 0.3001 & 9.79 & 6.71 & $4.77{\times}$ \\
 & On  & $5.405{\times}10^{-5}$ & 0.2543 & 7.63 & 6.71 & $4.77{\times}$ \\
\hline
Transform
 & $K=1$ & $5.471{\times}10^{-5}$ & 0.2684 & 8.36 & 6.71 & $4.77{\times}$ \\
 & $K=4$ & $5.405{\times}10^{-5}$ & 0.2543 & 7.63 & 6.71 & $4.77{\times}$ \\
\hline
Calibration
 & 4  & $5.548{\times}10^{-5}$ & 0.2579 & 7.47 & 6.71 & $4.77{\times}$ \\
 & 8  & $5.462{\times}10^{-5}$ & 0.2566 & 7.42 & 6.71 & $4.77{\times}$ \\
 & 16 & $5.423{\times}10^{-5}$ & 0.2546 & 7.93 & 6.71 & $4.77{\times}$ \\
 & 32 & $5.405{\times}10^{-5}$ & 0.2543 & 7.63 & 6.71 & $4.77{\times}$ \\
\hline
Statistics
 & 16-bit & $5.405{\times}10^{-5}$ & 0.2537 & 7.63 & 5.79 & $5.53{\times}$ \\
 & 32-bit & $5.405{\times}10^{-5}$ & 0.2543 & 7.63 & 6.71 & $4.77{\times}$ \\
\hline
Block rows
 & 64  & $5.380{\times}10^{-5}$ & 0.2548 & 7.55 & 6.74 & $4.74{\times}$ \\
 & 128 & $5.405{\times}10^{-5}$ & 0.2543 & 7.63 & 6.71 & $4.77{\times}$ \\
 & 256 & $5.422{\times}10^{-5}$ & 0.2586 & 7.71 & 6.69 & $4.78{\times}$ \\
\hline
Vector $(n,m)$
 & $(128,128)$ & $5.176{\times}10^{-5}$ & 0.2449 & 6.92 & 8.40 & $3.81{\times}$ \\
 & $(256,256)$ & $5.405{\times}10^{-5}$ & 0.2543 & 7.63 & 6.71 & $4.77{\times}$ \\
 & $(512,512)$ & $5.526{\times}10^{-5}$ & 0.2622 & 8.43 & 5.87 & $5.45{\times}$ \\
\hline
\end{tabular}
\endgroup%
}
\end{minipage}
\par\medskip

\par\medskip
\noindent\begin{minipage}{\columnwidth}
\centering
\captionof{table}{Llama-3-8B component ablations ($r=5$, joint scope).}
\label{tab:llama_ablation}
\resizebox{\columnwidth}{!}{%
\begin{tabular}{llrrrrr}
\hline
\textbf{Ablation} & \textbf{Setting} & \textbf{Param. MSE}$\downarrow$ & \textbf{Logit MSE}$\downarrow$ & \textbf{$\Delta$PPL (\%)}$\downarrow$ & \textbf{BPW}$\downarrow$ & \textbf{Comp.}$\uparrow$ \\
\hline
AAS & Off & $3.379\times10^{-6}$ & 1.0983 & 26.10 & 6.65 & $4.81\times$ \\
 & On & $6.867\times10^{-6}$ & 0.6920 & 13.81 & 6.66 & $4.81\times$ \\
\hline
Transform & $K=1$ & $6.944\times10^{-6}$ & 0.6994 & 14.04 & 6.66 & $4.81\times$ \\
 & $K=4$ & $6.867\times10^{-6}$ & 0.6920 & 13.81 & 6.66 & $4.81\times$ \\
\hline
Calibration & 4 & --- & 0.6804 & 14.45 & 6.66 & $4.81\times$ \\
 & 8 & --- & 0.6561 & 13.55 & 6.66 & $4.81\times$ \\
 & 16 & --- & 0.6769 & 13.80 & 6.66 & $4.81\times$ \\
 & 32 & --- & 0.6920 & 13.81 & 6.66 & $4.81\times$ \\
\hline
Statistics & 16-bit & --- & 0.6904 & 13.77 & 5.76 & $5.55\times$ \\
 & 32-bit & --- & 0.6920 & 13.81 & 6.66 & $4.81\times$ \\
\hline
Block rows & 64 & --- & 0.6730 & 14.00 & 6.70 & $4.77\times$ \\
 & 128 & --- & 0.6920 & 13.81 & 6.66 & $4.81\times$ \\
 & 256 & --- & 0.6447 & 14.48 & 6.63 & $4.83\times$ \\
\hline
Block batch & 1 & --- & 0.6918 & 13.84 & 6.66 & $4.81\times$ \\
 & 4 & --- & 0.6920 & 13.81 & 6.66 & $4.81\times$ \\
 & 8 & --- & 0.6918 & 13.84 & 6.66 & $4.81\times$ \\
\hline
\end{tabular}%
}
\end{minipage}
\par\medskip

\noindent\emph{Note:} Dashes indicate parameter-MSE values unavailable at sufficient precision in the supplied table; they are not zero.

The tables report all tested ablations; the main findings are summarized in Section~7.

\FloatBarrier
\section{Additional Joint-Scope Quantizer Benchmark}
\label{sec:additional_benchmark}
This benchmark compares scalar, residual vector, identity-sign, and orthogonal-sign quantizers on joint attention+MLP weights of GPT-2 Large and Llama-3-8B. It uses FP16 reconstructed dense weights, vector/block dimension 256, 32-bit sign statistics, a 4,096-token input (4,095 scored tokens), and $r\in\{1,2,3,5\}$. Scalar baselines use 2- or 4-bit residual quantization; residual VQ uses dimension 8, codebooks of size 16 or 256, and five iterations. Unlike the scope sweeps, this benchmark evaluates reconstructed dense weights, so their perplexities are not directly comparable.

\paragraph{Payload and runtime.}
Estimated bits per weight (BPW) equal eight times estimated payload bytes divided by the count of selected parameters. This excludes unquantized weights, checkpoint framing, and runtime storage. Dense-weight evaluation cannot establish compressed inference speedup; encoder timings are implementation-dependent. Compare payloads rather than equal residual depths.

\begin{figure}[!tbp]
\centering
\includegraphics[width=\columnwidth,height=0.31\textheight,keepaspectratio]{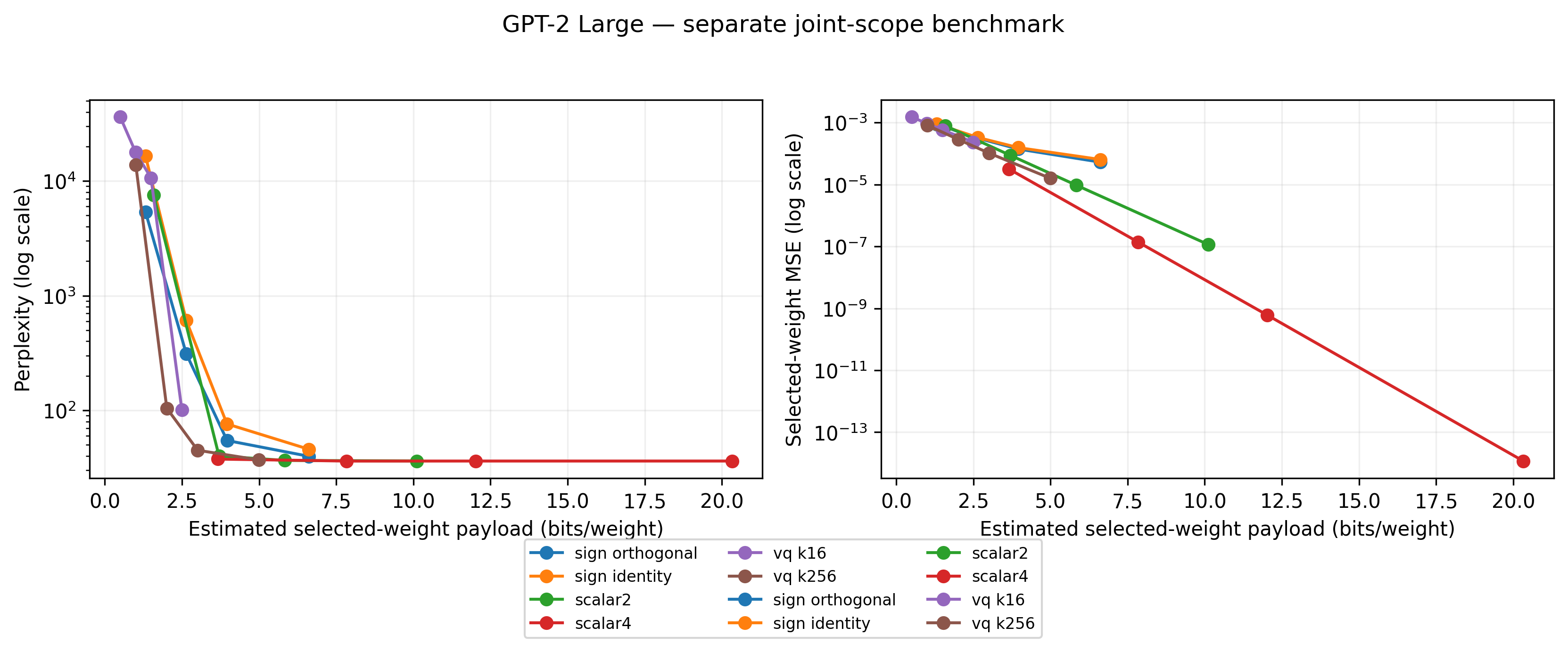}
\caption{GPT-2 Large joint-scope quantizer comparison.}
\label{fig:joint_benchmark_gpt2}
\end{figure}
\begin{figure}[!tbp]
\centering
\includegraphics[width=\columnwidth,height=0.31\textheight,keepaspectratio]{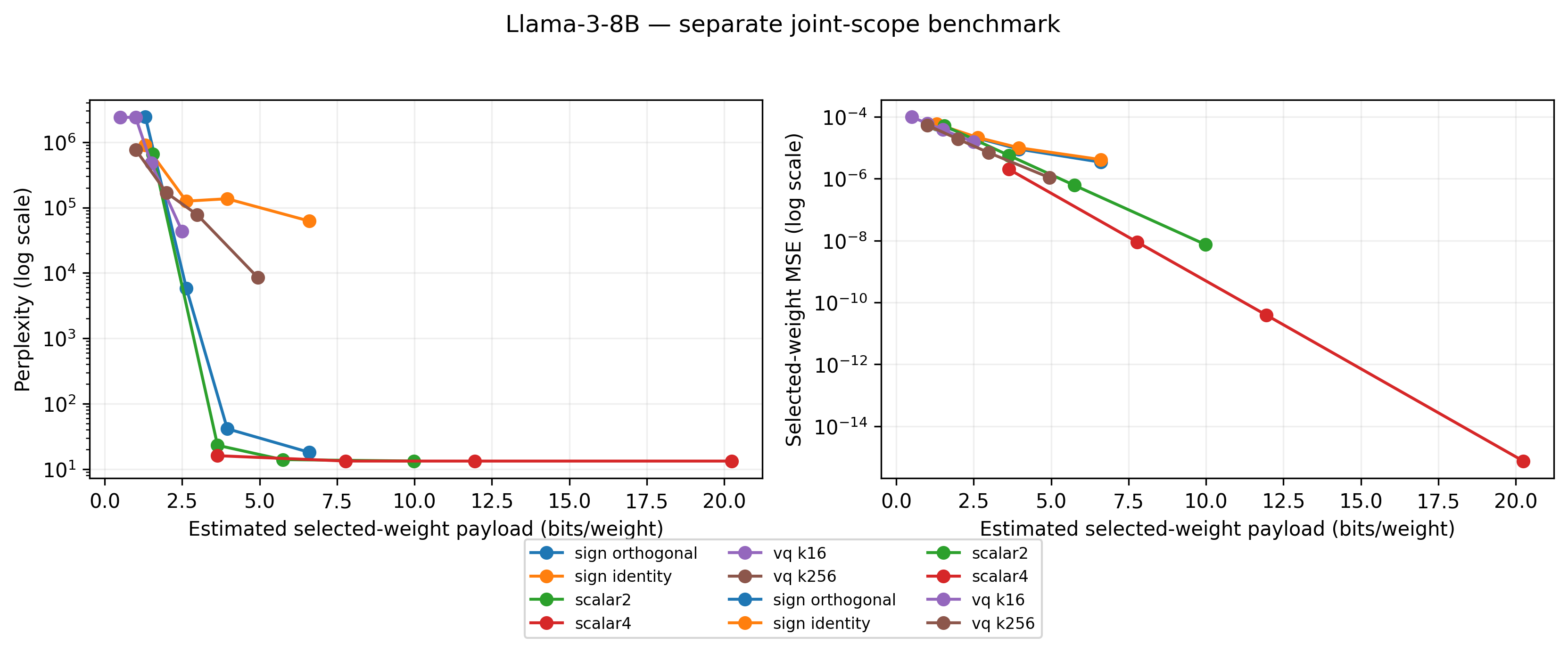}
\caption{Llama-3-8B joint-scope quantizer comparison.}
\label{fig:joint_benchmark_llama3}
\end{figure}
\paragraph{Progressive refinement and mixing.}
In Fig.~\ref{fig:joint_benchmark_gpt2}, GPT-2 Large orthogonal sign coding improves from PPL $5368.89$ and MSE $8.755\times10^{-4}$ at $r=1$ (1.319 BPW) to PPL $39.73$ and MSE $5.265\times10^{-5}$ at $r=5$ (6.618 BPW). Identity-sign coding at $r=5$ reaches PPL $45.81$ and MSE $6.428\times10^{-5}$ (6.611 BPW). In Fig.~\ref{fig:joint_benchmark_llama3}, Llama-3-8B orthogonal sign coding improves from PPL $2.425\times10^6$ and MSE $5.660\times10^{-5}$ at $r=1$ (1.313 BPW) to PPL $18.02$ and MSE $3.405\times10^{-6}$ at $r=5$ (6.611 BPW); identity-sign coding at $r=5$ gives PPL $62178.91$ and MSE $4.136\times10^{-6}$ (6.602 BPW). At $r=1$, however, orthogonal sign coding has worse Llama-3-8B perplexity than identity coding ($2.425\times10^6$ versus $9.03\times10^5$).

\paragraph{Scalar and vector baselines.}
On GPT-2 Large, scalar 4-bit at $r=1$ achieves PPL $37.59$ at $3.658$ BPW, and residual VQ ($K=256$, $r=5$) achieves $36.98$ at $4.995$ BPW; both outperform orthogonal sign coding at $r=5$ ($39.73$, $6.618$ BPW) on this rate--perplexity comparison. On Llama-3-8B, scalar 4-bit at $r=1$ yields PPL $16.05$ at $3.632$ BPW, versus orthogonal sign's $18.02$ at $6.611$ BPW. Residual VQ ($K=256$, $r=5$) yields PPL $8449$ at $4.948$ BPW, far worse than the sign codec but at a lower rate. The sign encoder is faster than the tested VQ implementations, but timings are not matched by rate or kernel. The scalar 1-bit entries have zero accepted residual stages, negligible estimated payload, and catastrophic perplexity; they do not constitute a functioning 1-bit baseline and are excluded from comparative plots. These results support progressive codebook-free coding and orthogonal mixing, but not universal rate--distortion or inference-speed superiority.

\FloatBarrier
\end{document}